\documentclass[conference]{IEEEtran}
\IEEEoverridecommandlockouts
\usepackage{cite}
\usepackage{amsmath,amssymb,amsfonts}
\usepackage{algorithmic}
\usepackage{graphicx}
\usepackage{textcomp}
\usepackage{xcolor}
\usepackage{algorithm}
\usepackage{algorithmic}
\usepackage{booktabs}
\usepackage{multirow}
\usepackage{array}
\usepackage{threeparttable}
\usepackage{graphicx} %
\usepackage{enumitem}
\usepackage{amsmath}
\usepackage{graphicx}
\usepackage{algorithm}
\usepackage{amsfonts}
\usepackage{algorithmic}
\usepackage{times}
\usepackage{helvet}
\usepackage{courier}
\usepackage{xcolor}
\usepackage{url}
\def\BibTeX{{\rm B\kern-.05em{\sc i\kern-.025em b}\kern-.08em
    T\kern-.1667em\lower.7ex\hbox{E}\kern-.125emX}}
\begin{document}

\title{FlashSign: Pose-Free Guidance for Efficient Sign Language Video Generation}

\author{
Liuzhou Zhang$^{1}$
Zeyu Zhang$^{2*}$~
Biao Wu$^{3}$~
Luyao Tang$^{4}$~
Zirui Song$^{5}$~
Hongyang He$^{6\dag}$~
Renda Han$^{7}$\\
Guangzhen Yao$^{8}$~
Huacan Wang$^{9}$~
Ronghao Chen$^{10}$~
Xiuying Chen$^{5}$~
Guan Huang$^{2}$~
Zheng Zhu$^{2}$\\
\small
$^1$HUST~
$^2$GigaAI~
$^3$UTS~
$^4$HKU~
$^5$MBZUAI~
$^6$Warwick~
$^7$TJU~
$^8$NNU~
$^9$UCAS~
$^{10}$UTHealth Houston~\\
\footnotesize
$^*$Project lead.
$^\dag$Corresponding author: hongyang.he@warwick.ac.uk.
\vspace{-2em}
}

\maketitle

\begin{abstract}
Sign language plays a crucial role in bridging communication gaps between the deaf and hard-of-hearing communities. However, existing sign language video generation models often rely on complex intermediate representations, which limits their flexibility and efficiency. In this work, we propose a novel pose-free framework for real-time sign language video generation. Our method eliminates the need for intermediate pose representations by directly mapping natural language text to sign language videos using a diffusion-based approach. We introduce two key innovations: (1) a pose-free generative model based on the a state-of-the-art diffusion backbone, which learns implicit text-to-gesture alignments without pose estimation, and (2) a Trainable Sliding Tile Attention (T-STA) mechanism that accelerates inference by exploiting spatio-temporal locality patterns. Unlike previous training-free sparsity approaches, T-STA integrates trainable sparsity into both training and inference, ensuring consistency and eliminating the train-test gap. This approach significantly reduces computational overhead while maintaining high generation quality, making real-time deployment feasible. Our method increases video generation speed by 3.07× without compromising video quality. Our contributions open new avenues for real-time, high-quality, pose-free sign language synthesis, with potential applications in inclusive communication tools for diverse communities.
Code:~\url{https://github.com/AIGeeksGroup/FlashSign}.
\end{abstract}

\section{Introduction}
Sign language is a vital means of communication for the deaf and hard-of-hearing community~\cite{WHO2022hearing}, but there is still a significant gap in accessible communication tools for this group~\cite{sincan2021chalearn,wei2023improving}. The development of sign language video generation systems holds great promise for bridging this gap by enabling real-time, text-to-sign video synthesis. Recent advances in deep generative models, particularly diffusion-based architectures, have shown considerable success in generating high-quality, temporally coherent videos. However, applying these models to sign language generation presents unique challenges, primarily due to the reliance on intermediate pose estimation~\cite{bangham2000virtual,kim2022sign} and the high computational cost associated with diffusion models~\cite{chen2023control,duarte2021how2sign,esser2023structure}, which hinders their real-time deployment.

\begin{figure}[t!]  %
    \centering
    \includegraphics[width=0.5\textwidth]{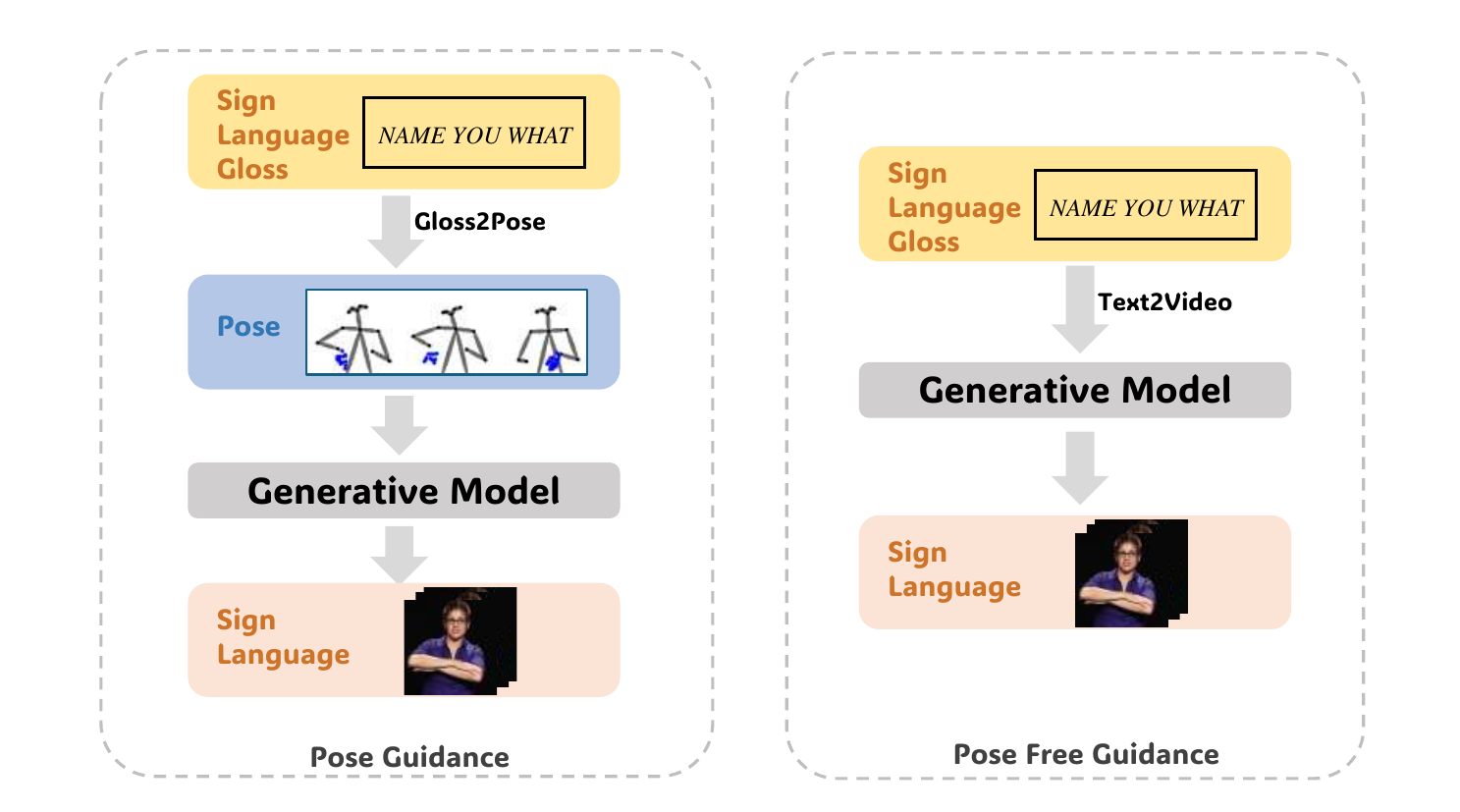}
    \caption{
        Comparison between \textbf{Pose Guidance} (left) and \textbf{Pose-Free Guidance} (right).
    }
    \label{fig:pfg}
\end{figure}

Unlike conventional video generation tasks, where the focus is on overall visual quality, sign language video generation places greater emphasis on the smoothness and accuracy of motion, particularly in localized regions such as the hands and arms.~\cite{zhang2025towards,lei2024comprehensive} This distinction requires specialized attention to inference efficiency and fine-grained temporal modeling. Existing sign language video generation methods typically follow a two-stage pipeline: text-to-pose generation followed by pose-to-video synthesis. Although this approach can yield reasonable results, it suffers from two main limitations. First, errors in pose estimation can propagate throughout the pipeline, leading to visual artifacts and poor semantic alignment. Second, pose-based methods often fail to generalize to real-world scenarios, particularly in zero-shot or pose-absent situations such as blind-deaf communication.

Moreover, diffusion-based models, although capable of producing high-quality outputs~\cite{sun2024hunyuanlargeopensourcemoemodel,kong2025hunyuanvideosystematicframeworklarge}, are computationally expensive and often impractical for real-time applications because of the long inference times required to generate video sequences~\cite{zhang2025fast,xi2025sparse,dong2412flex}. This further exacerbates the challenge of enabling real-time interaction in sign language communication.

Notably, sign language videos typically feature static backgrounds and relatively fixed upper body postures, with the meaningful content concentrated in the hand regions. This spatial redundancy presents an opportunity to optimize the computational resources required for synthesis~\cite{pan2024t,he2024zipvl}. Building on this observation, we introduce a key innovation: a unified Trainable Sliding Tile Attention (T-STA) mechanism that is jointly optimized during training and seamlessly applied during inference. Unlike previous training-free sparsity methods that often suffer from mismatch between training and inference time attention patterns, T-STA enforces sparsity in a consistent, learnable manner across both stages. By dynamically partitioning attention maps into spatial-temporal tiles and training the model to focus on gesture-intensive regions, T-STA achieves fine-grained control over computation while preserving critical motion dynamics. This alignment between training and inference ensures stable performance and avoids the generalization gap observed in heuristic sparse attention techniques. As a result, T-STA delivers a substantial reduction in inference latency without compromising video fidelity.

To overcome the limitations of traditional pose-based methods and the high computational demands of diffusion models~\cite{wang2024loong,zhang2024sageattention}, we propose a novel framework that achieves near-real-time video generation speeds. The first core innovation is a pose-free generative paradigm, built upon the latest video diffusion model~\cite{wan2025wanopenadvancedlargescale}, which directly maps natural language text to sign language videos without relying on intermediate pose representations. This method bypasses the traditional pose estimation step, learning implicit text-to-gesture alignments, and demonstrating strong robustness in zero-shot and blind-deaf communication scenarios. The second key innovation is the introduction of the T-STA mechanism, designed to optimize computational efficiency while maintaining high generation quality. By partitioning the attention map into spatial-temporal tiles and applying dynamic tiling strategies, T-STA adapts to both fine-grained temporal motion and low-frequency spatial structures, resulting in a substantial reduction in inference time.

Together, these innovations enable the generation of high-quality, pose-free sign language videos at significantly reduced inference latency, bringing real-time sign language synthesis within reach. Our method sets a new standard for generating robust, high-fidelity sign language videos while addressing the practical challenges of real-time deployment.

To summarize, our contributions are as follows:
\begin{itemize}
    \item We propose a pose-free sign language video generation framework, enabling end-to-end synthesis from natural language to sign language video without the need for intermediate pose estimation.
    \item We introduce a unified training-inference sliding tile attention (STA) mechanism that achieves efficient and consistent computation, enabling faster inference without compromising video quality. Unlike prior training-free sparsity approaches, T-STA maintains alignment between training and inference sparsity patterns, ensuring stable performance and eliminating generalization gaps.
    \item We conducted comprehensive experiments showing that our method increases generation speed by 3.07× without compromising video quality, while delivering outstanding performance in both generation quality and inference speed.
\end{itemize}

\begin{figure*}[t!]
    \centering
    \includegraphics[width=0.9\textwidth]{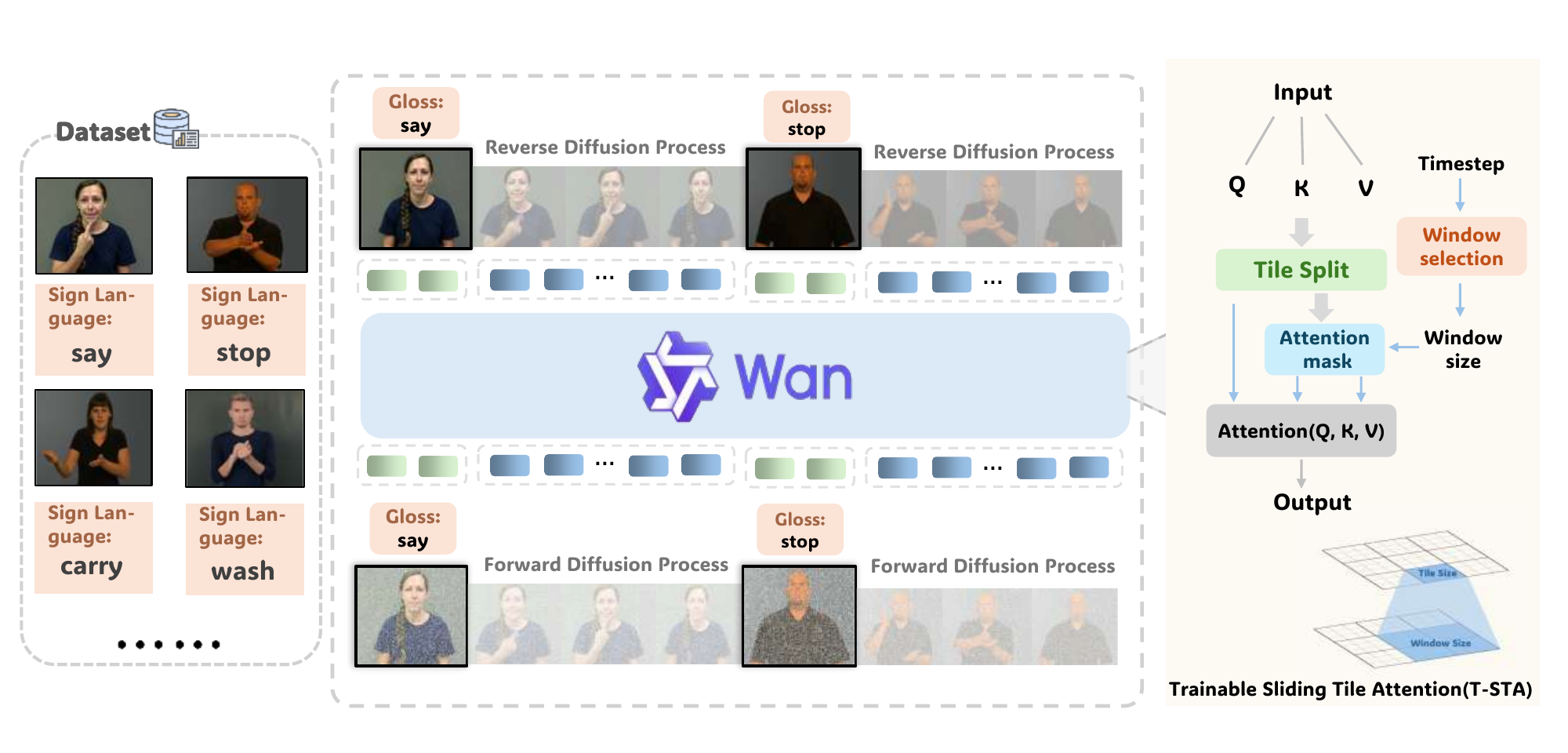}
    \caption{
        Overall pipeline of the FlashSign.
    }
    \label{fig:flashsign_overview}
\end{figure*}

\section{The Proposed Method}

\subsection{Overall Pipeline}

We propose an end-to-end sign language video generation framework that directly maps text to video, eliminating the need for intermediate pose representations. Traditional pipelines follow a pose-guided strategy, decomposing the task into two stages:
\[
\mathbf{P} = f_{\text{text2pose}}(\mathbf{T}), \quad \mathbf{V} = f_{\text{pose2vid}}(\mathbf{P}),
\]
where \( \mathbf{T} \in \mathbb{R}^{L \times d_T} \) is the tokenized input text, \( \mathbf{P} \in \mathbb{R}^{T \times J \times 3} \) denotes the pose sequence across \( T \) frames and \( J \) joints, and \( \mathbf{V} \in \mathbb{R}^{T \times H \times W \times 3} \) is the resulting RGB video. Although this decomposition facilitates motion planning, it suffers from cascading errors, supervision bottlenecks, and increased inference cost.

In contrast, our approach adopts a pose-free guidance mechanism based on conditional diffusion:
\[
\mathbf{V} = f_{\text{text2vid}}(\mathbf{T}),
\]
where \( f_{\text{text2vid}} \) learns to synthesize videos directly from language prompts, without relying on explicit motion supervision. This design streamlines training, improves generalization in low-resource settings, and simplifies deployment in real-world scenarios.

To further improve efficiency, we introduce T-STA (Trainable Sparse Tile Attention), a lightweight attention module that replaces standard full self-attention with sparse, learnable tile patterns in both spatial and temporal dimensions. Unlike fixed-window sparsity, T-STA dynamically adapts to content structure and denoising timestep, enabling efficient and expressive attention computation. This makes our model substantially faster during inference while maintaining high video quality, facilitating real-time sign language synthesis.

\subsection{End-to-End Pose Free Guidance}

In recent sign language video generation frameworks, a typical approach involves converting text into a sequence of human poses, followed by rendering corresponding video frames. Although this two-stage pipeline simplifies motion representation, it is inherently limited by pose estimation errors, which can lead to artifacts and poor expressiveness. In contrast, we propose a pose-free video generation framework that directly synthesizes videos from text, eliminating the need for pose information. Our method relies on a latent diffusion model that operates within a continuous latent space learned via a 3D Variational Autoencoder (VAE).

Given an input video \( v \in \mathbb{R}^{T \times H \times W \times C} \), the 3D VAE encoder maps this video to a compact latent representation \( z \in \mathbb{R}^{T' \times H' \times W' \times C'} \). The latent variable \( z \) is assumed to follow a normal distribution:
\[
z \sim \mathcal{N}(\mu, \sigma^2), \quad \mu, \sigma^2 = \mathcal{E}(v),
\]
where \( \mu \) and \( \sigma^2 \) are the mean and variance parameters produced by the encoder. This continuous latent space captures the essential spatio-temporal structure of the video while significantly reducing its dimensionality. The compactness of this representation allows for efficient generative modeling while retaining important motion and appearance features.

To guide the generation process, we employ a pre-trained UmT5~\cite{chung2023unimax} text encoder that maps the input text to a shared text-visual embedding space. This embedding \( z_x \in \mathbb{R}^d \) acts as a semantic guidance signal during the generation process. At each timestep \( t \), the model conditions the denoising process on both the noised latent \( z_t \) and the text embedding \( z_x \). The role of the text embedding is to ensure that the generated video aligns with the semantic content described in the input text.

During generation, we simulate a forward diffusion process by progressively adding Gaussian noise to the latent representation. Given the clean latent \( z_0 \), we corrupt it at each timestep \( t \) according to:
\[
z_t = \sqrt{\bar{\alpha}_t} z_0 + \sqrt{1 - \bar{\alpha}_t} \epsilon, \quad \epsilon \sim \mathcal{N}(0, I),
\]
where \( \bar{\alpha}_t \) is a schedule controlling the noise level at each timestep. This process generates noisy versions of the latent representation, which are then denoised in reverse.

The denoising process is modeled by a Video Diffusion Transformer (DiT), which applies 3D full attention to capture both spatial and temporal dependencies in the latent space. At each timestep, the model predicts the noise residual \( \epsilon \) conditioned on the noised latent \( z_t \) and the text embedding \( z_x \). The denoising objective is to minimize the error between the predicted and true noise:
\[
\mathcal{L}_{\text{diff}} = \mathbb{E}_{z_0, t, \epsilon \sim \mathcal{N}(0, I)} \left[ \| \epsilon - \epsilon_\theta(z_t, t, z_x) \|^2 \right].
\]
The model progressively refines the latent representation, removing noise and aligning it with the semantic guidance provided by the text.

Once the latent representation \( z_0 \) is denoised, we decode it back into the video space using the VAE decoder \( \mathcal{D} \):
\[
\hat{v} = \mathcal{D}(z_0).
\]
This final decoding step reconstructs high-resolution video frames that are visually coherent and semantically consistent with the input text. By removing the need for pose supervision, our approach not only simplifies the pipeline but also improves the flexibility and generalization capabilities, particularly in zero-shot settings, where the model is tasked with generating videos from unseen text descriptions.

\subsection{Trainable Sliding Tile Sparse Acceleration}

We propose a \textit{Trainable Sliding Tile Attention} (T-STA) mechanism that substantially enhances the traditional Sliding Tile Attention (STA) by leveraging trainable, dynamic attention windows which adapt to the noise timestep in the diffusion process. Unlike conventional methods dependent on static, pre-set windows, T-STA flexibly adjusts its attention window sizes during both training and inference according to the current noise level. This timestep-dependent adaptive design not only improves model performance but also ensures computational efficiency critical for real-time video synthesis.

\vspace{0.5em}
\noindent\textbf{Tile Partitioning:} The input video sequence is first partitioned into tiles along spatiotemporal dimensions with preset tile sizes \((t_{\mathrm{tile}}, h_{\mathrm{tile}}, w_{\mathrm{tile}})\):
\[
i = \operatorname{tile\_idx}(p) \in \{1, \ldots, N_{\mathrm{tile}}\},
\]
where \(p\) denotes the position index in the sequence, and \(N_{\mathrm{tile}}\) is the total number of tiles. This partition preserves local spatial structure and drastically reduces computational complexity.

\vspace{0.5em}
\noindent\textbf{Piecewise Window Size:} For each timestep \(t \in \{1, \ldots, T\}\), the window size is adaptively determined by segment index \(s \in \{1,\ldots,L\}\):
$$
(T_w(t), H_w(t), W_w(t)) =
(T_w^{(s)}, H_w^{(s)}, W_w^{(s)}), 
$$
enabling the model to utilize larger windows during early noise timesteps to capture coarse spatiotemporal dependencies while gradually shrinking the window to focus on fine details at later timesteps, thus balancing accuracy and efficiency.

\vspace{0.5em}
\noindent\textbf{Mask Construction for Local and Cross-modal Attention:} To dynamically restrict attention computation within relevant local neighborhoods, we construct a block mask based on the coordinates of query and key tiles in the spatiotemporal space. For each query tile, a sliding window is defined around its position using half-window radii derived from the current timestep’s adaptive window size. To ensure the window does not exceed the spatial or temporal bounds of the input sequence, the query tile coordinates are clamped within valid ranges. The block mask then includes only those key tiles located within this clamped, localized window across time, height, and width dimensions. This design significantly reduces computational cost by focusing the attention on contextually important nearby regions, while maintaining the flexibility to adjust the window size according to noise levels in the diffusion process.

In addition, to enable rich multimodal interactions, we introduce a cross-modal mask that governs the attention between visual and textual modalities. The mask supports three key interaction patterns: image tokens attending to other image tokens to preserve visual coherence; image tokens attending to text tokens to incorporate linguistic guidance into visual representations; and text tokens attending to both image and text tokens to fully leverage the combined semantic context. This structured masking facilitates comprehensive cross-modal information exchange, enhancing the model's ability to fuse language and vision effectively for tasks such as sign language video generation and other multimodal applications.

\vspace{0.5em}
\noindent\textbf{Attention Computation:} At timestep \(t\), the masked attention scores and output are computed as:
\[
\begin{aligned}
\widetilde{\mathbf{A}}_t &= \frac{Q_t K_t^\top}{\sqrt{d}} + \log M_t, \\
\mathbf{A}_t &= \mathrm{softmax}(\widetilde{\mathbf{A}}_t), \\
\mathrm{Output}_t &= \mathbf{A}_t V_t,
\end{aligned}
\]
where \(Q_t, K_t, V_t \in \mathbb{R}^{B \times H \times S \times d}\) denote query, key, and value matrices respectively, and \(\log M_t\) implements masking by assigning \(-\infty\) to invalid positions.

\vspace{0.5em}
\noindent\textbf{Loss Function and Window Size Optimization:} During diffusion model training, the objective minimizes the noise prediction error:
\[
\mathcal{L} = \mathbb{E}_{\mathbf{x}_0, t \sim \mathrm{Unif}[1,T], \epsilon \sim \mathcal{N}(0, I)} \left\| \epsilon - \epsilon_\theta(\mathbf{x}_t, t, \mathbf{A}_t) \right\|^2,
\]
where the noise estimator \(\epsilon_\theta\) is conditioned on both the noisy input \(\mathbf{x}_t\), timestep \(t\), and dynamic attention mask \(\mathbf{A}_t\). Crucially, the window size parameters \(\{T_w^{(s)}, H_w^{(s)}, W_w^{(s)}\}\) are treated as trainable variables, optimized via gradients during training to adaptively balance window configurations across noise levels, thereby improving generation quality and computational efficiency.

In summary, T-STA employs \textit{Tile Partitioning} and \textit{Piecewise Window Sizes} to locally model spatiotemporal dependencies, while dynamically computed \textit{Block Masks} ensure focused attention within relevant regions. Combined with cross-modal attention interactions and window size optimization through loss-driven training, the proposed approach achieves real-time, high-quality performance with significant computational savings in sign language video generation and other multimodal tasks.

\begin{figure}[t!]  %
    \centering
    \includegraphics[width=0.5\textwidth]{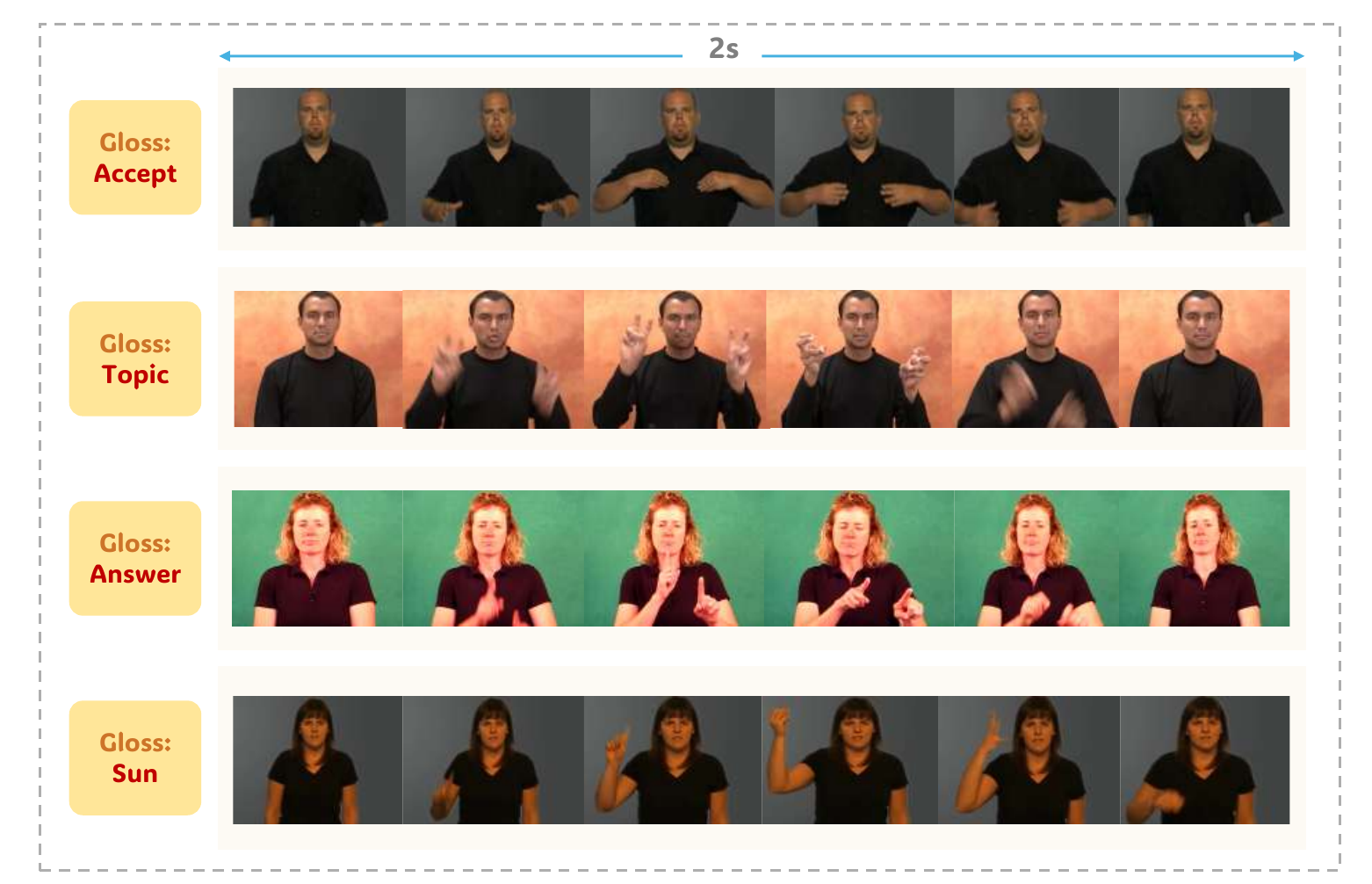}
    \caption{
        A demo frame showcasing sign language video generation.
    }
    \label{fig:heatmap_freq}
\end{figure}

\section{Experiments}

\subsection{Dataset}
Our sign language video generation framework utilizes the large-scale WLASL (Word-Level American Sign Language)~\cite{LI_2020_WACV} dataset as the foundation for experiments. The dataset comprises 21,083 videos featuring 119 signers performing 2,000 distinct ASL glosses (words). To ensure robustness against inter-signer variations, each gloss is performed by at least three different signers. With approximately 14 hours of video content, this dataset represents the largest vocabulary resource currently available for ASL research. The videos have an average duration of 2.41 seconds, and each gloss contains an average of 10.5 samples. The inclusion of dialectal variations further enhances the dataset's real-world applicability.

\subsection{Evaluation Metric}

To rigorously evaluate the visual quality, perceptual fidelity, and temporal coherence of generated sign language videos, we adopt the following metrics:

\begin{itemize}
    \item \textbf{Peak Signal-to-Noise Ratio (PSNR):} 
    Measures pixel-level reconstruction accuracy between generated and ground-truth videos. Higher PSNR (in dB) indicates lower distortion, though it may not fully capture perceptual quality.
    
    \item \textbf{Structural Similarity Index (SSIM):} 
    Assesses perceptual similarity by comparing luminance, contrast, and structural patterns. SSIM values range from 0 to 1, with higher values denoting better preservation of spatial features.
    
    \item \textbf{Learned Perceptual Image Patch Similarity (LPIPS):} 
    Quantifies perceptual differences using deep features (e.g., VGG or AlexNet). Lower LPIPS scores (closer to 0) indicate higher perceptual similarity to ground truth.
    
    \item \textbf{Fréchet Video Distance (FVD):} 
    Evaluates temporal coherence and motion realism by comparing feature statistics of generated and real video sequences. Lower FVD reflects more natural dynamics.
\end{itemize}

These metrics jointly address pixel fidelity (PSNR), structural integrity (SSIM), human perception (LPIPS), and motion plausibility (FVD), ensuring a comprehensive assessment of sign language generation.

\subsection{Comparison with Baselines}
As shown in Table~\ref{tab:baseline}, our framework significantly outperforms current mainstream sparse attention variants (SageAttention, SpargeAtten, SparseVideoGen, SignGen, and STA) across five key metrics. Specifically, our method achieves the best score of 453 on FVD (Fréchet Video Distance), which measures video dynamic quality, representing an 8.1\% improvement over the pose-guided SignGen method (493) and 8\%-24\% improvements over other baseline models. This advantage is particularly crucial for sign language video generation as it directly reflects the temporal coherence and naturalness of generated motions.

In terms of perceptual metrics, we achieve new state-of-the-art results with 0.07 LPIPS (Learned Perceptual Image Patch Similarity) and 0.83 SSIM (Structural Similarity Index Measure), demonstrating significant advantages in preserving details such as hand shapes and facial expressions. Notably, while SignGen slightly outperforms in PSNR (Peak Signal-to-Noise Ratio) with 27.72, this metric has limited reference value for assessing perceptual quality in dynamic sign language videos.

More importantly, our end-to-end training scheme completely eliminates dependence on intermediate pose estimation, which not only avoids error accumulation problems but also significantly improves the model's generalization capability for complex sign language vocabulary. The experimental results demonstrate that this integrated training-inference paradigm can surpass traditional pose-guided methods, providing a more reliable solution for practical sign language generation applications.

\begin{table}[t]
\caption{Performance comparison with baseline methods on key metrics for sign language video generation. Lower values are better for FVD and LPIPS, while higher values are better for PSNR and SSIM.}
\label{tab:baseline}
\centering
\resizebox{\columnwidth}{!}{%
\begin{tabular}{lcccc}
\toprule
Method &FVD$\downarrow$ & PSNR$\uparrow$ & SSIM$\uparrow$ & LPIPS$\downarrow$ \\
\midrule
SageAttention &501  & 24.33 & 0.82 & 0.15 \\
SpargeAtten &597  & 21.38 & 0.81 & 0.21 \\
SparseVideoGen &523  & 23.52 & 0.80 & 0.17 \\
SignGen &493 &\textbf{27.72} &0.81 & 0.08 \\
STA &519  & 22.65 & 0.82 & 0.19 \\
\textbf{Ours}&\textbf{453}   & 25.74 & \textbf{0.83} & \textbf{0.07} \\
\bottomrule
\end{tabular}}
\end{table}

\subsection{Ablation Studies}
In this section, we conduct an ablation study to assess the impact of various design choices in our model, including the influence of the attention window size, tile size, and denoising steps. These experiments allow us to identify the optimal hyperparameters for balancing generation quality and inference speed.
\begin{figure}[t!]
    \centering
    \includegraphics[width=\linewidth]{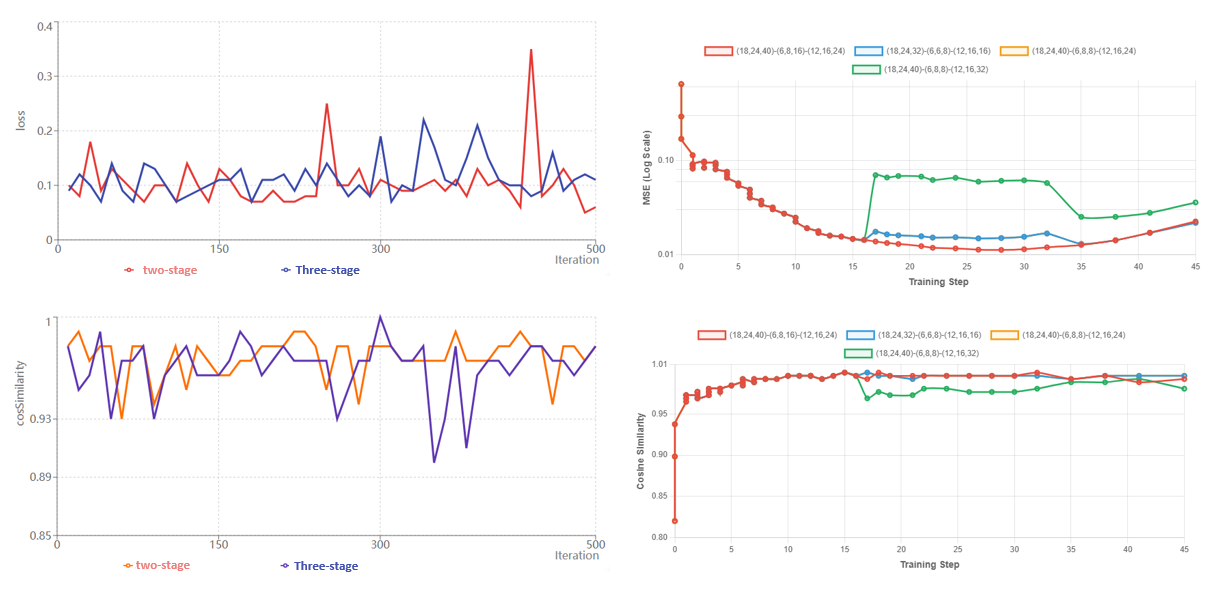}
    \caption{
        Performance comparison between two-stage windows and three-stage windows during the training process.
    }
    \label{fig:mse}
\end{figure}

Our window configuration analysis reveals interesting patterns in training dynamics. As shown in Figure~\ref{fig:mse}, the three-stage configuration with intermediate window sizes demonstrates clear advantages in training stability. The progressive transition from $(18,24,32)$ to $(12,16,24)$ and finally to $(12,16,16)$ creates a smoother learning curve, particularly in the depth dimension where it bridges the gap between the initial larger receptive field (32) and the final smaller one (16). This design not only accelerates convergence but also leads to more robust feature learning throughout the training process.

The tile size analysis yields equally insightful results (Table~\ref{tab:tile_sizes}). We observe a consistent positive correlation between tile dimensions and model performance across all metrics. Notably, the largest tile size ($24\times32\times32$) achieves optimal performance with PSNR 25.74, SSIM 0.83, and LPIPS 0.07. While this configuration increases memory requirements, the performance gains justify this trade-off for most applications. Interestingly, the improvement from $(6,8,8)$ to $(12,16,16)$ shows diminishing returns, suggesting that practitioners might consider intermediate tile sizes when computational resources are constrained.

These ablation studies collectively demonstrate that careful tuning of window configurations and tile sizes significantly impacts model performance. The three-stage window design offers superior training dynamics, while larger tile sizes consistently improve generation quality. The smile-shaped convergence curves in our supplementary materials further validate these design choices, showing smooth optimization trajectories without abrupt changes or instabilities, providing valuable guidance for practitioners implementing similar architectures.

\begin{table}[h]
\centering
\caption{Impact of tile size variations on Wan2.1 1.3B model performance. We assess different combinations of temporal (t), height (h), and width (w) dimensions.}
\label{tab:tile_sizes}
\resizebox{0.8\columnwidth}{!}{%
\begin{tabular}{ccccccc}
\toprule
\multicolumn{3}{c}{Tile Size} & \multicolumn{3}{c}{Metrics} \\
\cmidrule(lr){1-3} \cmidrule(lr){4-6}
{t} & {h} & {w} & {PSNR $\uparrow$} & {SSIM $\uparrow$} & {LPIPS $\downarrow$} \\
\midrule
3 & 4 & 4 & 19.87 & 0.77 & 0.16 \\
6 & 8 & 8 & 24.12 & 0.82 & 0.10 \\
12 & 16 & 16 & 24.45 & 0.81 & 0.09 \\
24 & 32 & 32 & 25.74 & 0.83 & 0.07 \\
\bottomrule
\end{tabular}}
\end{table}

\section{Limitations and Future Work}

While our pose-free guidance framework demonstrates promising results in generating isolated sign language words, its capacity to synthesize coherent multi-sign sequences remains limited. The absence of explicit pose supervision restricts the model’s ability to preserve fine-grained spatial-temporal consistency across longer utterances, which is essential for sentence-level generation. As the semantic complexity and temporal length of the target sequence increase, we anticipate that stronger guidance, such as structured pose priors or hierarchical constraints, will become necessary to ensure grammaticality and fluid motion transitions. Future work will explore hybrid strategies that integrate sparse pose information to support sentence-level synthesis without fully reverting to pose-dependent pipelines.

Although our approach shows great promise, several areas remain for future exploration. First, expanding the model to support a wider variety of sign languages and dialects will enhance its global applicability. Second, optimizing the framework for resources-constrained devices will make it more accessible in everyday environments. Lastly, incorporating non-manual signals, such as facial expressions, could further enrich the expressiveness of the generated videos.

\section{Conclusions}

In this paper, we presented a pose-free sign language video generation framework that directly maps natural language text to high-quality sign language videos, overcoming the limitations of traditional pose-guided methods. By introducing a dynamic sliding tile attention (T-STA) mechanism, we achieved significant improvements in both video quality and inference speed, enabling real-time generation. Overall, our work provides a new foundation for real-time, pose-free sign language synthesis and paves the way for more inclusive communication tools.

\appendix

\section*{Related Works}

\subsection{Sign Language Generation}
Sign language generation (SLG) has been extensively studied as a means to facilitate communication between hearing individuals and those who are deaf or hard of hearing, with a particular emphasis on sign language avatars. Early systems such as Tessa~\cite{cox2002tessa}, DictaSign~\cite{efthimiou2012dicta}, and Sign3D~\cite{gibet2016interactive} employed 3D animated models to simulate both manual and non-manual components of sign language. To enhance realism, later works incorporated motion capture technologies, improving motion fidelity at the expense of increased data acquisition costs. More recently, deep learning-based pose generation methods have gained traction. ~\cite{saunders2020progressive} proposed a Progressive Transformer to convert spoken language into 3D sign pose sequences, and transformer-based models such as Text2Gloss2Pose~\cite{saunders2020progressive} and Gloss2Pose~\cite{saunders2022signing} mapped gloss sequences into continuous spatio-temporal pose representations. The Text2Gloss2Pose2Video framework~\cite{stoll2018sign} extended this approach by generating gloss sequences from text, transforming them into skeleton poses, and conditioning generative models for video synthesis. ~\cite{saunders2022signing} introduced a frame selection network and a pose-guided human synthesis model to enhance temporal consistency and visual realism. More recent approaches combined large language models and motion matching to improve the naturalness and expressiveness of generated videos~\cite{karras2019style}, while ~\cite{kipp2011sign} proposed \textit{SignGen}, an end-to-end diffusion model that directly generates sign language videos from text and integrates non-manual signals through motion-aware generation and multimodal condition fusion. These methods primarily adopt a two-stage pipeline: mapping text or gloss into pose sequences followed by video synthesis, which has led to substantial progress in sign-language avatar generation.

\subsection{Efficient Video Generation}
With the growing success of Diffusion Transformers (DiTs)~\cite{peebles2023scalable} in high-fidelity video generation~\cite{duan2026liveworld,wu2026light4d,zhang2025egolcd,zhang2025blockvid,shi2025presentagent,wang2025drivegen3d,luo2025univid}, their considerable computational cost has become a major obstacle to scalable deployment. To mitigate this limitation, recent studies have focused on leveraging sparsity patterns to improve inference efficiency without sacrificing generation quality~\cite{zhang2024sageattention2,zhangsageattention2,zhang2025spargeattn,zhangspargeattn,liu2025fpsattention}. Sparse VideoGen~\cite{xi2025sparse} proposes a training-free black-box acceleration framework that dynamically identifies spatial-temporal sparsity patterns in attention heads through online profiling. This is paired with optimized tensor layout transformations and custom kernels to enhance runtime efficiency. Efficient-vDiT~\cite{ding2025efficientvditefficientvideodiffusion} addresses structural redundancy by pruning unnecessary connections in 3D attention modules and incorporating consistency distillation to streamline the sampling process, achieving faster inference with minimal quality degradation. AdaSpa~\cite{xia2025trainingfreeadaptivesparseattention} introduces a block-structured hierarchical sparsity method that combines dynamic pattern search with a fused LogSumExp caching mechanism to improve sparse attention indexing efficiency. Complementarily, Sliding Tile Attention (STA)~\cite{Zhang2025FastVG} reformulates the attention mechanism by introducing a sliding tile-based local attention structure, which eliminates redundant global computations by exploiting the locality of attention in spatio-temporal neighborhoods. Collectively, these approaches systematically reduce computational redundancy through attention compression, architectural sparsification, and operator-level optimization, offering practical pathways toward efficient and scalable video generation.

\section*{Inference Performance Evaluation}
We conducted comprehensive performance testing on the Wan2.1-based sign language video generation model. The test platform utilized a tensor-parallel computing cluster consisting of four NVIDIA H20 GPUs (TP=4 configuration). The evaluation generated 5-second sign language videos (9:16 aspect ratio, 480p resolution, 16fps frame rate, totaling 81 frames) through a 50-step diffusion sampling process (sample\_steps=50) with classifier-free guidance scale (CFG Scale) set to 5.0. The video generation process was executed using synchronized multi-GPU parallel computing (TP-4 mode). To ensure reliable test results, we performed three independent experiments and reported the average inference latency metrics.

\begin{table}[t]
\centering
\caption{Inference latency comparison of different models. The baseline models are WanX2.1-1.3B (271s) and WanX2.1-14B (1301s). Our method achieves speedups of 2.32× and 3.07× on the 1.3B and 14B models respectively, significantly outperforming other attention optimization approaches.}
\label{tab:latency}
\resizebox{0.8\columnwidth}{!}{%
\begin{tabular}{lcc}
\toprule
Method & Latency$\downarrow$  & Speedup$\uparrow$  \\
\midrule
WanX2.1-1.3B  & 271s & 1.00x  \\
\midrule
SpargeAtten  &205s	&1.32x \\
SageAttention  & 141s	&1.91x  \\
SparseVideoGen  &152s	&1.78x  \\
STA  &143s	&1.89x \\
\textbf{Ours}  & 117s & 2.32x  \\
\midrule
WanX2.1-14B  & 1301s & 1.00x  \\
\midrule
SpargeAtten  & 734s & 1.77x  \\
SageAttention  & 646s	&2.01x  \\
SparseVideoGen & 613s & 2.12x \\
STA  & 548s & 2.37x  \\
\textbf{Ours}  & 423s	 &3.07x \\
\bottomrule
\end{tabular}}
\end{table}

Experimental results demonstrate that various attention optimization methods (SparseAtten~\cite{tay2020sparse}, SageAttention~\cite{zhang2024sageattn}, SparseVideoGen~\cite{xi2025sparse} and STA) can significantly reduce inference latency across models of different scales. For the 1.3B parameter model, our method achieved the best performance with a 2.32× speedup, reducing latency from 271s to 117s, outperforming other approaches (SparseAtten: 1.32×, SageAttention: 1.91×, SparseVideoGen: 1.78×, STA: 1.89×).

This performance advantage becomes even more pronounced in the 14B parameter model, where our approach achieves a remarkable 3.07× speedup (423s vs. baseline 1301s), significantly surpassing other methods (SparseAtten: 1.77×, SageAttention: 2.01×, SparseVideoGen: 2.12×, STA: 2.37×). These results not only confirm our method's superior acceleration capability across model scales but also demonstrate its excellent scalability for larger models. The complete latency comparison is shown in Table~\ref{tab:latency}.

\bibliographystyle{IEEEbib}
\bibliography{icme2026references}

\end{document}